\documentclass{article}

\PassOptionsToPackage{square,sort,comma,numbers}{natbib}

\usepackage[preprint]{neurips_2024}

\usepackage[utf8]{inputenc} 
\usepackage[T1]{fontenc}    
\usepackage{hyperref}       
\usepackage{url}            
\usepackage{booktabs}       
\usepackage{amsfonts}       
\usepackage{nicefrac}       
\usepackage{microtype}      
\usepackage{xcolor}         
\usepackage[pdftex]{graphicx}
\usepackage{subcaption}
\usepackage{multirow}
\usepackage{multicol}

\hypersetup{
	colorlinks,
	linkcolor={red!80!black},
	citecolor={blue!80!black},
	urlcolor={blue!80!black}
}

\title{One-shot Training for Video Object Segmentation}

\author{%
  Baiyu Chen \\
  Key Laboratory of Intelligent Informatics for Safety \& Emergency of Zhejiang Province\\
  Wenzhou University\\
  Wenzhou, 325035, China \\
  \texttt{baiyuchen@stu.wzu.edu.cn} \\
  \And
  Sixian Chan \\
  The College of Computer Science and Technology \\
  Zhejiang University of Technology\\
  Hangzhou, 310023, \\
  \texttt{sxchan@zjut.edu.cn} \\
  \AND
  Xiaoqin Zhang\thanks{Corresponding author.} \\
  Key Laboratory of Intelligent Informatics for Safety \& Emergency of Zhejiang Province \\
  Wenzhou University \\
  Wenzhou, 325035, China \\
  \texttt{zhangxiaoqinnan@gmail.com} \\
}

\begin{document}

\maketitle

\begin{abstract}
Video Object Segmentation (VOS) aims to track objects across frames in a video and segment them based on the initial annotated frame of the target objects. Previous VOS works typically rely on fully annotated videos for training. However, acquiring fully annotated training videos for VOS is labor-intensive and time-consuming. Meanwhile, self-supervised VOS methods have attempted to build VOS systems through correspondence learning and label propagation. Still, the absence of mask priors harms their robustness to complex scenarios, and the label propagation paradigm makes them impractical in terms of efficiency. To address these issues, we propose, for the first time, a general one-shot training framework for VOS, requiring only a single labeled frame per training video and applicable to a majority of state-of-the-art VOS networks. Specifically, our algorithm consists of: i) Inferring object masks time-forward based on the initial labeled frame. ii) Reconstructing the initial object mask time-backward using the masks from step i). Through this bi-directional training, a satisfactory VOS network can be obtained. Notably, our approach is extremely simple and can be employed end-to-end. Finally, our approach uses a single labeled frame of YouTube-VOS and DAVIS datasets to achieve comparable results to those trained on fully labeled datasets. The code will be released.
\end{abstract}

\section{Introduction} \label{intro}

Video Object Segmentation (VOS) has recently made remarkable and expeditious progresses~\cite{Xmem,Cutie}. Most works concentrates on training effective VOS networks using fully annotated VOS datasets, such as YouTube-VOS~\cite{youtube-vos} and  DAVIS~\cite{davis2016,davis2017}. However, training these networks necessitates both spatially dense annotations, i.e., pixel-level, and temporally dense annotations, which incurs substantial costs in dataset acquisition. For example, the DAVIS dataset comprises 60 videos with an average of 70 annotated frames per video, while the YouTube-VOS dataset includes over 3000 videos with dense labeling on every fifth frame. Consequently, it is important to develop label-efficient approaches to mitigate the extensive annotation burden of VOS networks. 

Recent efforts in developing unsupervised VOS models~\cite{Mast,CRW,DUL} have demonstrated promising outcomes. These approaches typically regard the VOS task as a combination of unsupervised correspondence learning and non-parametric mask warping. While the learned correspondences facilitate object correlation across frames, predicting object masks through non-parametric mask warping proves inefficient and vulnerable to occlusion and drift. Conversely, a method similar to ours aims to train VOS networks effectively in a semi-supervised manner~\cite{two-shot-vos}. However, its implementation differs significantly from ours; it utilizes two annotated frames per video and involves multiple training phases, whereas our approach requires only a single labeled frame per training video and can be simply achieved through an end-to-end way.

\begin{figure}[!t]
    \centering
    \includegraphics[width=\linewidth]{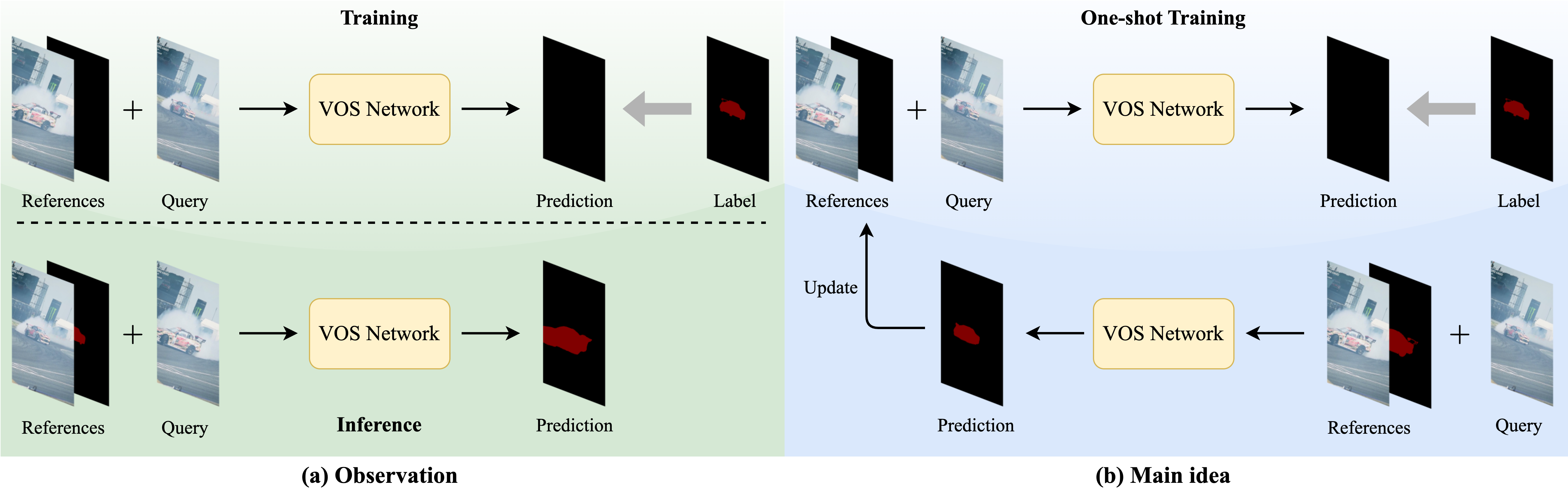}
    \caption{The main concept illustration. \textbf{(a)} We found that existing video object segmentation networks trained from a noisy reference mask (empty or all-black) can predict the rough mask in a video sequence. \textbf{(b)} To utilizing this property, we regard the rough prediction as the noisy reference to build a feedback loop, and propose a straightforward One-shot Training framework for video object segmentation, which exhibits great label-efficiency and generalization.}
    \label{fig:key_concept}
\end{figure}

In this work, we demonstrate the feasibility of training an effective VOS network using a single labeled frame per video. For the first time, we introduce an end-to-end one-shot VOS training framework that can be broadly applied to most state-of-the-art VOS networks. The central concept is based on the observation that coarse object masks can emerge from a noisy reference mask, as depicted in Fig.~\ref{fig:key_concept}~(a). Specifically, by training a VOS network with only one annotated frame and the noisy reference mask of the initial frame, the network can approximately predict the masks of target objects in subsequent frames. Leveraging this property, we iteratively process through the VOS network by treating its predictions as noisy inputs. This bi-directional training approach enables the VOS network to be effectively trained in a feedback loop, where the initial noisy references yield meaningful predictions that subsequently improve input quality. This concept is illustrated in Fig.~\ref{fig:key_concept}~(b).

Our contributions can be summarized as follows:
\begin{itemize}
\item We demonstrate the feasibility of training an effective VOS network using a single labeled frame per video.
\item We introduce, for the first time, an end-to-end single-frame supervised training framework that efficiently trains a VOS network with just one labeled frame per video, showing strong generalization across various state-of-the-art VOS networks.
\item Our approach achieves performance on par with fully labeled data counterparts on the YouTube-VOS~\cite{youtube-vos} and DAVIS~\cite{davis2016,davis2017} benchmarks by utilizing only a single labeled frame per video.
\end{itemize}

\section{Related Works}

\paragraph{Fully-supervised VOS.} Early methods addressing the VOS challenge employed an online fine-tuning approach~\cite{one-shot-vos,static_images,OAVOS,wotemporal,fastandrobust}, leading to impractically long inference times. Recent developments have shifted towards offline training of VOS networks~\cite{STM,Xmem,aot,xmempp}, utilizing propagation or memory matching mechanisms to achieve significant improvements in both accuracy and efficiency. Propagation-based approaches~\cite{state-aware-tracker,dyenet,RGMP} utilize the most recent past mask to sequentially predict subsequent masks. Memory matching-based methods~\cite{STM,STCN,Xmem,Cutie} enhance the use of past masks by constructing a memory bank to store features from previous frames. In this study, we focus primarily on memory matching-based methods due to their simplicity and efficiency. Specifically, STCN~\cite{STCN} effectively models space-time correspondences for VOS, while XMem~\cite{Xmem}, an extension of STCN, incorporates multiple types of memory for improved performance. Cutie~\cite{Cutie} introduces top-down object-level memory reading to better address challenging scenarios. Despite the promising results demonstrated by memory matching-based approaches in VOS, their training paradigm is hindered by the high cost of manual labeling.

\paragraph{Self-supervised VOS.} Self-supervised VOS aims to develop a VOS system without requiring labeled data, typically decomposing the VOS task into unsupervised correspondence learning and non-parametric mask warping. During training, visual correspondences across frames are learned using methods such as photometric reconstruction~\cite{colorizing,Mast}, cycle-consistency~\cite{cycle-consist,CRW}, or contrastive matching~\cite{DUL,rethink_corr_learn}. During inference, past masks are warped to generate subsequent masks based on affinities derived from the learned correspondences. However, constructing a VOS system without any mask priors proves insufficient for tracking complex motions in practice, and relying on mask warping during inference is highly inefficient.

\paragraph{Semi-supervised VOS.} In contrast, two-shot VOS~\cite{two-shot-vos} employs a multi-phase training process that starts with model-specific pre-training to generate pseudo labels~\cite{pseduo_label,lee2013pseudo} for subsequent model-agnostic training, utilizing up to two labeled frames. The major distinctions between our approach and this method are: 1) our approach utilizes only a single labeled frame during training; 2) our one-shot training paradigm is entirely model-agnostic; and 3) our approach can be implemented in an end-to-end manner.

\section{One-shot Training} \label{sec:method}

\begin{figure}[h!]
    \centering
    \includegraphics[width=\linewidth]{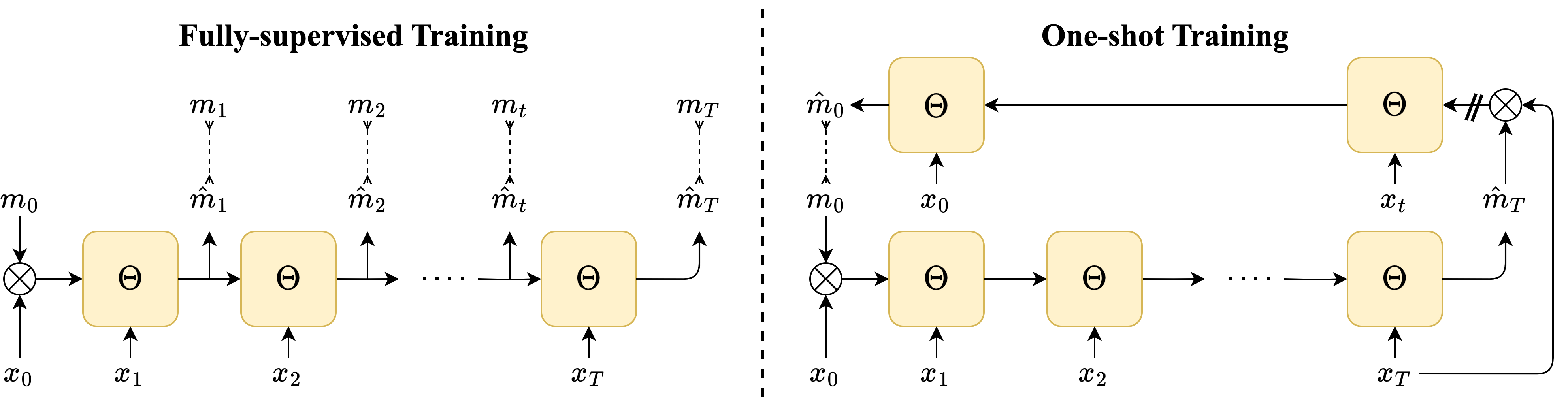}
    \caption{\textbf{Left}: The traditional fully-supervised training paradigm for VOS networks. At each time, object masks predicted through a VOS network $\Theta$ are aligned with the corresponding labels, updating the network $\Theta$. \textbf{Right}: Our proposed One-shot Training framework for VOS networks. We first use a VOS network $\Theta$ to infer masks time-forward from the initial labeled frame, but without matching the intermidiate predictions. Then, we reconstruct the initial object mask from the last prediction mask.}
    \label{fig:comparison_training}
\end{figure}

\subsection{Preliminary}

In this paper, we particularly focus on offline-learning VOS methods, which can be categorized into propagation-based and memory matching-based approaches. Since memory matching-based approaches can be considered a generalization~\cite{STM} of propagation-based methods, we primarily discuss the fully-supervised training paradigm in the context of memory matching-based VOS networks.

As depicted on the left of Fig.~\ref{fig:comparison_training}, given a training video clip in which each frame $x_t$ is associated with a labeled mask $m_t$, the VOS networks are trained to align the predictions with the labeled masks at each frame. At time $T$, a VOS method maintains a memory bank storing features extracted from the past frames $\mathcal{X}_{t<T}=\{x_0,x_1,...,x_{T-1}\}$ and masks $\mathcal{M}_{t<T}=\{m_0,\hat{m}_1,...,\hat{m}_{T-1}\}$. By computing the affinities of $\mathcal{X}_{t<T}$ and the query frame $x_T$, a VOS network $\Theta$ aggregates the past features to obtain the prediction mask $m_T$ at time $T$. For general purpose, we simplify the formulation at time $T$ to a function that takes a triplet:
\begin{equation}
    \hat{m}_{T}=\Theta(x_{T},\mathcal{X}_{t<T},\mathcal{M}_{t<T}).
    \label{eqn:prev_vos}
\end{equation}

For a training video clip, predictions $\hat{m}_t$ at arbitrary time are aligned with the ground truth $m_t$ by optimizing a model-specific loss, which is typically a cross entropy loss~\cite{STM,STCN}, or a combination of cross entropy loss and dice loss~\cite{Xmem}, or more complicated losses~\cite{Cutie}. And we generally abstract it as $\Delta$ and express it as following:
\begin{equation}
    \mathcal{L}=-\Delta(\hat{m}_t, m_t).
    \label{eqn:loss-func}
\end{equation}

\subsection{Overview}

In contrast to conventional fully-supervised training, our goal is to train a VOS network on videos which only the annotation of the initial frame is available. Our approach involves a bi-directional prediction, allowing the VOS network to be properly optimized only on the initial labeled frame. In the \emph{time-forward inference}, we obtain an object mask following Eq.~(\ref{eqn:prev_vos}). In the \emph{time-backward reconstruction}, by using the object mask inferred from forward pass, we predict the initial mask time-backward. Finally, we update the VOS network by aligning the prediction initial mask and the initial ground truth mask. Our approach is simple yet effective, illustrated on the right of Fig.~\ref{fig:comparison_training}.

\subsection{Time-forward Inference}
\label{sec:forward-inference}

The formulation of forward process is similar to the fully-supervised VOS training, with the distinction that loss computation is not performed during time-forward inference due to the absence of the subsequent frames annotations. For a training video $\mathcal{X}=\{x_0,x_1,...,x_T\}$ and its initial ground truth mask $m_0$, we employ a VOS network to predict sequentially over $T$-step, yielding $\hat{m}_T$. The procedure is identical to Eq.~(\ref{eqn:prev_vos}). It is worth noting that this process can be operated without gradient since there are not labels for frames to match masks obtained from the time-forward inference calculation. 

\begin{figure}[t]
    \centering
    \includegraphics[width=\linewidth]{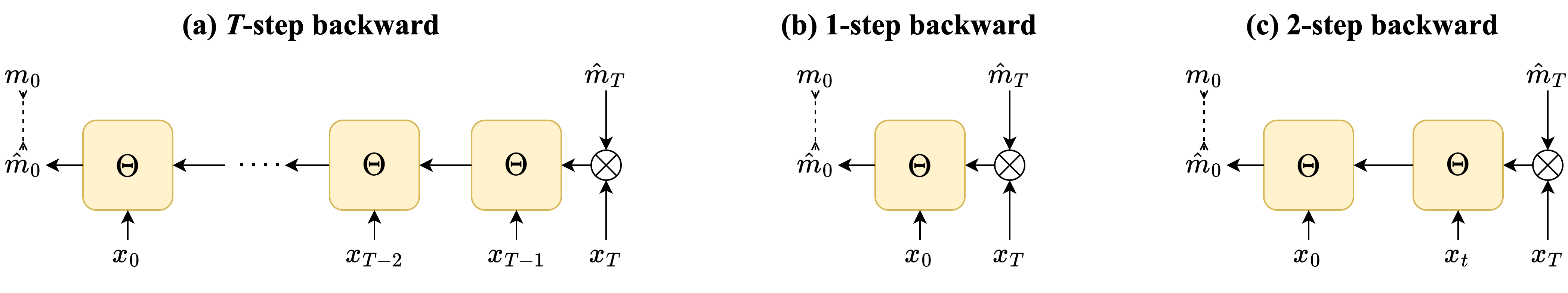}
    \caption{\textbf{(a) \emph{T}-step backward}: simlpy reversing frames in time, and go through the VOS network $\Theta$ similar to time-forward inference. \textbf{(b) 1-step backward}: regarding the initial frame adjacent to the current frame, directly go through $\Theta$ once to reconstruct the initial mask. \textbf{(c) 2-step backward}: randomly sampling another frame as the only intermediate frame between the current frame and the initial frame, go through $\Theta$ twice to predict the initial mask.}
    \label{fig:backward_illustration}
\end{figure}

\subsection{Time-backward Reconstruction}
\label{sec:backward-reconstruction}

As soon as obtaining a prediction from forward inference at time $T$, we are able to reversely traverse back to time 0, \emph{i.e.}, the initial time, and predict the initial object mask. From the perspective of network optimization, we consider three backward strategies: \emph{T-step backward}, \emph{1-step backward} and \emph{2-step backward}. Fig.~\ref{fig:backward_illustration} illustrates the key differences among these three patterns.

\paragraph{\emph{T}-step backward.} One naive solution is to simply flip the training video sequence in a time-reversed order, and autoregressively predict the object masks, which can be termed as \emph{$T$-step backward}. Specifically, by flipping $\mathcal{X}$, we get a reversed sequence $\mathcal{X}^\prime=\{x_T,x_{T-1},...,x_0\}$. And the mask at arbitrary time $T_k$ can be produced as following:
\begin{equation}
    \hat{m}^\prime_{T_k}=\Theta(x_{T_k},\mathcal{X}^\prime_{t>T_k},\mathcal{X}^\prime_{t>T_k}),
    \label{eqn:backward-vos}
\end{equation}
\begin{equation}
    \mathcal{X}_{t>T_k}^\prime=\{x_T,x_{T-1},...,x_{T_k-1}\} \quad \mathcal{M}^\prime_{t>T_k}=\{m_T,m_{T-1},...,m_{T_k-1}\},
    \label{eqn:backward-mem}
\end{equation}
where the right upper prime symbol indicates that masks are predicted time-backward. 

We empirically found that the \emph{T-step backward} demonstrates the optimal results when VOS networks have been pretrained on synthetic datasets. However, it will predict $T$ masks while only the label of the initial frame is available, which is tricky to optimize when training from scratch. In Sec.~\ref{sec:analysis}, we will further discuss the role of different backward patterns as well as their impacts on One-shot Training.

\paragraph{1-step backward.} Another idea is to ignore frames between the initial frame and the $T^\texttt{th}$ frame, \emph{i.e.,} regarding they are adjacent. In this case, $\mathcal{X}^\prime=\{x_T,x_0\}$ and the initial mask $\hat{m}^\prime_0$ can be directly obtained by invoking Eq.~(\ref{eqn:backward-vos})(\ref{eqn:backward-mem}) once, hence 1-step. 

Although \emph{1-step backward} from $T^\texttt{th}$ to the initial frame well addresses the single labeled frame dilemma, it fails to work with VOS networks equipped with recurrent neural networks (RNNs), such as XMem~\cite{Xmem} and Cutie~\cite{Cutie}, which require at least 2~steps to be fully updated.

\paragraph{2-step backward.} Hence, a good trade-off between them is sampling one more frame between the the first frame and the $T^\texttt{th}$ frame to form a \emph{2-step backward} paradigm. Concretely, frame $x_{T_1}$ is randomly sampled from the video clip excluding the initial frame and the $T^{th}$ frame. Next, by sorting them in a time-reversed order, we end up with a set of frames like:
\begin{equation}
    \mathcal{X}^\prime=\{x_T,x_{T_1},x_0\},
\end{equation}
where $T>T_1>0$. Consequently, the initial mask $\hat{m}^\prime_0$ can be reconstructed following Eq.~(\ref{eqn:backward-vos})(\ref{eqn:backward-mem}).

After acquiring $\hat{m}^\prime_0$ with one of strategies above, we then compute a model-specific loss akin to Eq.~(\ref{eqn:loss-func}) to update the VOS network $\Theta$:
\begin{equation}
    \mathcal{L}=-\Delta(\hat{m}^\prime_0,m_0).
\end{equation}

\section{Experiments} 
\label{sec:exps}

We conduct experiments on the DAVIS 2016 validation set~\cite{davis2016}, DAVIS 2017 validation set~\cite{davis2017}  and YouTube-VOS 2018 validation set~\cite{youtube-vos}. To demonstrate the effectiveness and generalization of our approach, we employ One-shot Training on the state-of-the-art VOS networks: STCN~\cite{STCN}, XMem~\cite{Xmem} and Cutie~\cite{Cutie}, and compare results in contrast to their counterpart trained in the fully-supervised manner.

\subsection{Implementation Details} \label{sec:implementation}

We configure most the model-specific hyperparameters following the instructions released from their official codebase~\cite{STCN,Xmem,Cutie}. For each training video, $n$ temporally ordered frames (including the first frame) are sampled to form a training sequence, and the temporal distance between arbitrary two consecutive frames is limited to the maximum of $D$. $D$ is further maintained by a curriculum learning schedule. $n$ is set to 3, 8, and 8 for STCN, XMem and Cutie, respectively. The schedule of $D$ also varies from different VOS models. For example, the schedule for STCN is $[5,10,15,20,25,5]$ after $[0\%,0.1\%,0.2\%,0.3\%,0.4\%,0.8\%]$ training iterations correspondingly. To ensure fair comparison, we adopt the VOS networks pretrained on synthetic video datasets transformed from static image datasets, which only one label available per video. We also present results without pretrained on static image datasets for reference. 

Models are trained on the concatenation of YouTube-VOS 2018 and DAVIS 2017 training sets with two NVIDIA GPUs using PyTorch framework, and batch size is set to 4~, 8~and 16~correspond to STCN, XMem and Cutie. Besides, STCN uses an Adam optimizer with a weight decay of 1e-7. For XMem and Cutie, they utilize an AdamW optimizer with a weight decay of 0.05 and 0.001,~respectively. The learning rates of them follow a step learning rate decay schedule with a decay ratio of 0.1, and the corresponding initial learning rates are 1e-5, 1e-5, 1e-4. The complete configuration of hyperparameters can be found in the released code.

\subsection{Evaluation}

Being consistent with prior VOS works, we report results in region similarity, contour accuracy and their average for DAVIS 2016 and DAVIS 2017 benchmarks. For YouTube-VOS benchmark, we report $\mathcal{J}$ and $\mathcal{F}$ for both seen and unseen categories and the average score $\mathcal{G}$ using the official evaluation server.

\begin{table}[ht]
    \caption{Video object segmentation results on DAVIS 2016 and 2017 validation set. By merely using $1.4\%$ labels (1 annotated frame per training sequence) of DAVIS dataset, our approach achieves comparable results to fully-supervised ones (using $100\%$ labels).}
    \label{tab:davis_res}
    \centering
    \begin{tabular}{lccccccc}
        \toprule
        \multirow{2}{*}{Method} & \multirow{2}{*}{Labels}   & \multicolumn{3}{c}{DAVIS 2016 \emph{val}} & \multicolumn{3}{c}{DAVIS 2017 \emph{val}} \\
                                                            \cmidrule(lr){3-5} \cmidrule(lr){6-8}
                                &                            & $\mathcal{J}\&\mathcal{F}$ & $\mathcal{J}$ & $\mathcal{F}$ & $\mathcal{J}\&\mathcal{F}$ & $\mathcal{J}$ & $\mathcal{F}$\\
        \midrule
        OSMN~\cite{osmn} & 100\% & 73.5 & 74.0 & 72.9 & 54.8 & 52.5 & 57.1 \\
        FEELVOS~\cite{feelvos} & 100\% & 81.7 & 81.1 & 82.2 & 71.2 & 69.1 & 74.0 \\
        RGMP~\cite{RGMP} & 100\% & 81.8 & 81.5 & 82.0 & 66.7 & 64.8 & 68.6 \\
        Track-Seg~\cite{trackseg} & 100\% & 83.1 & 82.6 & 83.6 & 72.3 & 68.6 & 76.0 \\
        STM~\cite{STM} & 100\% & 89.3 & 88.7 & 89.9 & 81.8 & 78.2 & 84.3 \\
        CFBI~\cite{CFBI} & 100\% & 89.4 & 88.3 & 90.5 & 81.9 & 79.1 & 84.3 \\
        RDE-VOS~\cite{rdevos} & 100\% & 91.1 & 89.7 & 92.5 & 84.2 & 80.8 & 87.5 \\
        \midrule
        STCN~\cite{STCN} & 100\% & 91.6 & 90.8 & 92.5 & 85.4 & 82.2 & 88.6 \\
        1-shot STCN (Ours) & \textbf{1.4}\% & 89.8 & 88.8 & 90.8 & 81.7 & 78.6 & 84.8 \\
        \midrule
        XMem~\cite{Xmem} & 100\% & 91.5 & 90.4 & 92.7 & 86.2 & 82.9 & 89.5 \\
        1-shot XMem (Ours) & \textbf{1.4}\% & 90.1 & 89.7 & 90.6 & 82.6 & 79.8 & 85.4\\
        \midrule
        Cutie~\cite{Cutie} & 100\% & 89.2 & 88.0 & 90.4 & 88.8 & 85.4 & 92.3 \\
        1-shot Cutie (Ours) & \textbf{1.4}\% & 90.3 & 89.6 & 91.1 & 84.1 & 80.8 & 87.3 \\
        \bottomrule
    \end{tabular}
\end{table}

\begin{figure}[h!]
    \centering
    \includegraphics[width=\linewidth]{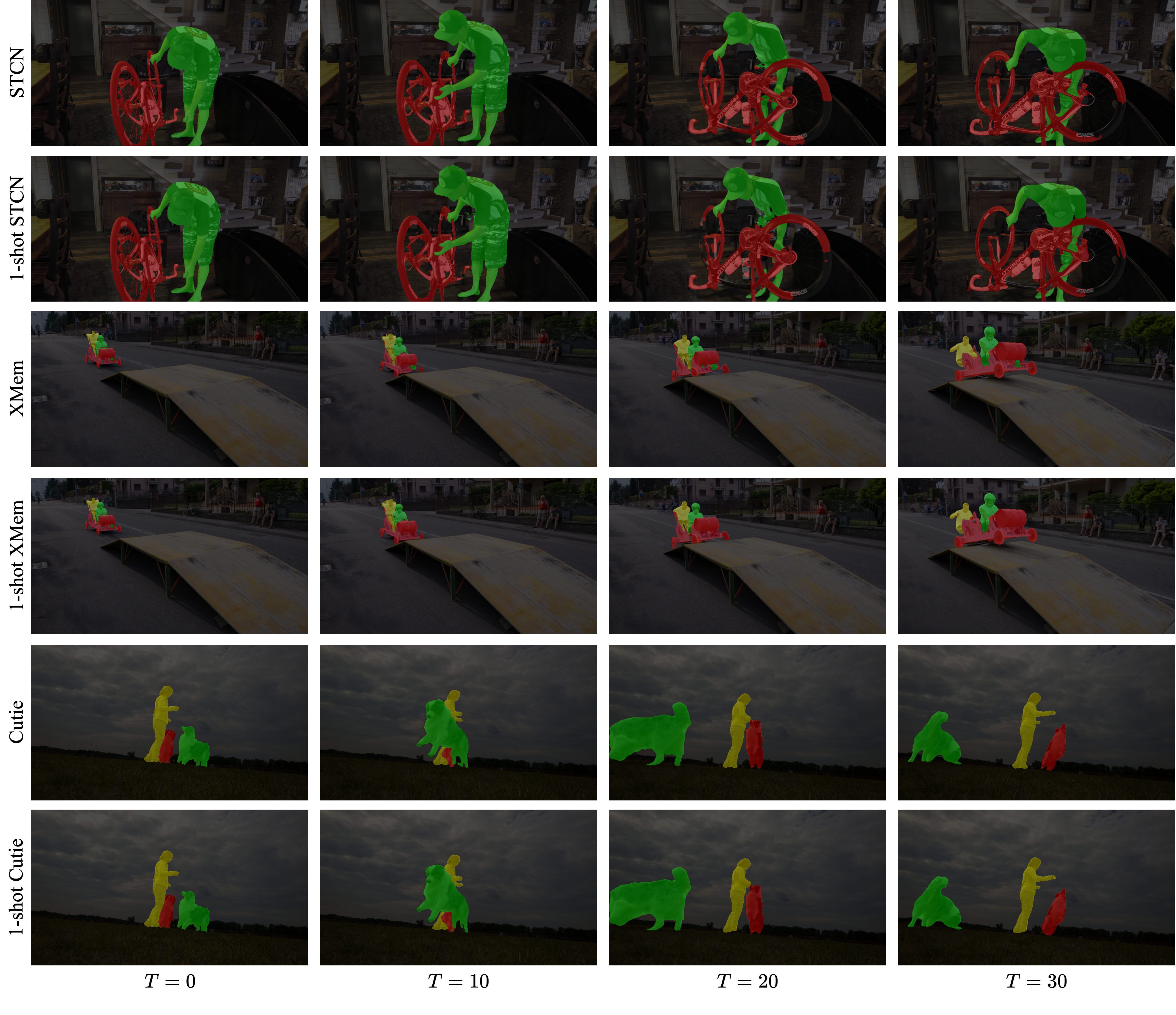}
    \caption{Qualitative comparisons of STCN~\cite{STCN}, XMem~\cite{Xmem} and Cutie~\cite{Cutie} trained with one-shot training and fully-supervised training on DAVIS 2017 \emph{val}. The first pair and third pair comparisons demonstrate that our approach achieves competitive performances in contrast to the fully-supervised. The second comparison on XMem shows that the fully-supervised model is more sensitive to details.}
    \label{fig:d17-vis}
\end{figure}

\subsection{Main Results}
\label{sec:main_res}

\paragraph{Comparisons on DAVIS benchmarks.} DAVIS 2016 \emph{val} is a single-object VOS benchmark, consists of 20 sequences. And DAVIS 2017 is a multiple-object extension to DAVIS 2016. In Table.~\ref{tab:davis_res}, we compare the video object segmentation results of training with our One-shot Training approach and training with fully-supervised method on three state-of-the-art models, STCN~\cite{STCN}, XMem~\cite{Xmem} and Cutie~\cite{Cutie}. As shown in Table.~\ref{tab:davis_res}, our approach achieves comparable results by using only $1.4\%$ labeled data of DAVIS datasets, \emph{i.e.}, one single annotated frame per training video. Moreover, our approach even outperforms the fully-supervised Cutie~\cite{Cutie} on the DAVIS 2016 benchmark, demonstrating the significant label-efficiency of our approach.

\begin{table}[htbp]
    \caption{Video object segmentation results on YouTube-VOS 2018 and 2019 validation set. By merely using $3.7\%$ labels (1 annotated frame per training sequence) of YouTube-VOS dataset, our approach shows promising results compared to the exisitng fully-supervised VOS methods.}
    \label{tab:ytvos_res}
    \centering
    \begin{tabular}{l@{\hspace{2mm}}cc@{\hspace{3mm}}c@{\hspace{3mm}}c@{\hspace{3mm}}c@{\hspace{3mm}}cc@{\hspace{3mm}}c@{\hspace{3mm}}c@{\hspace{3mm}}c@{\hspace{3mm}}c}
        \toprule
        \multirow{2}{*}{Method} & \multirow{2}{*}{Labels}   & \multicolumn{5}{c}{YouTube-VOS 2018 \emph{val}} & \multicolumn{5}{c}{YouTube-VOS 2019 \emph{val}} \\
                                                            \cmidrule(lr){3-7} \cmidrule(lr){8-12}
                             & & $\mathcal{G}$ & $\mathcal{J}_S$ & $\mathcal{F}_S$ & $\mathcal{J}_U$ & $\mathcal{F}_U$ & $\mathcal{G}$ & $\mathcal{J}_S$ & $\mathcal{F}_S$ & $\mathcal{J}_U$ & $\mathcal{F}_U$\\
        \midrule
        OSMN~\cite{osmn} & 100\% & 51.2 & 60.0 & 60.1 & 40.6 & 44.0 &-&-&-&-&- \\
        RGMP~\cite{RGMP} & 100\% & 53.8 & 59.5 & - & 45.2 & -&-&-&-&-&- \\
        Track-Seg~\cite{trackseg} & 100\% & 63.6 & 67.1 & 70.2 & 55.3 & 61.7 &-&-&-&-&-\\
        STM~\cite{STM} & 100\% & 79.4 & 79.7 & 84.2 & 72.8 & 80.9 & - & - & - & - & - \\
        CFBI~\cite{CFBI} & 100\% & 81.4 & 81.1 & 85.8 & 75.3 & 83.4 & 81.0 & 80.6 & 85.1 & 75.2 & 83.0 \\
        RDE-VOS~\cite{rdevos} & 100\% & - & - & - & - & - & 81.9 & 81.1 & 85.5 & 76.2 & 84.8 \\
        \midrule
        STCN~\cite{STCN} & 100\% & 83.0 & 82.0 & 86.5 & 77.8 & 85.8 & 82.7 & 81.2 & 85.4 & 78.2 & 86.0 \\
        1-shot STCN (Ours) & \textbf{3.7}\% & 79.3 & 77.6 & 74.8 & 81.4 & 83.4 & 78.8 & 76.8 & 74.8 & 80.3 & 83.1 \\
        \midrule
        XMem~\cite{Xmem} & 100\% & 85.5 & 84.4 & 89.1 & 80.0 & 88.3 & 85.3 & 84.0 & 88.2 & 80.4 & 88.4 \\
        1-shot XMem (Ours) & \textbf{3.7}\% & 79.6 & 79.9 & 72.5 & 84.5 & 81.6 & 79.3 & 79.2 & 72.9 & 83.5 & 81.6 \\
        \midrule
        Cutie~\cite{Cutie} & 100\% & 86.1 & 85.8 & 90.5 & 80.0 & 88.0 & 86.1 & 85.5 & 90.0 & 80.6 & 88.3 \\
        1-shot Cutie (Ours) & \textbf{3.7}\% & 79.8 & 79.8 & 73.2 & 84.3 & 81.8 &  79.3 & 79.2  & 73.2 & 83.3 & 81.3  \\
        \bottomrule
    \end{tabular}
\end{table}

\begin{figure}[h!]
    \centering
    \includegraphics[width=\linewidth]{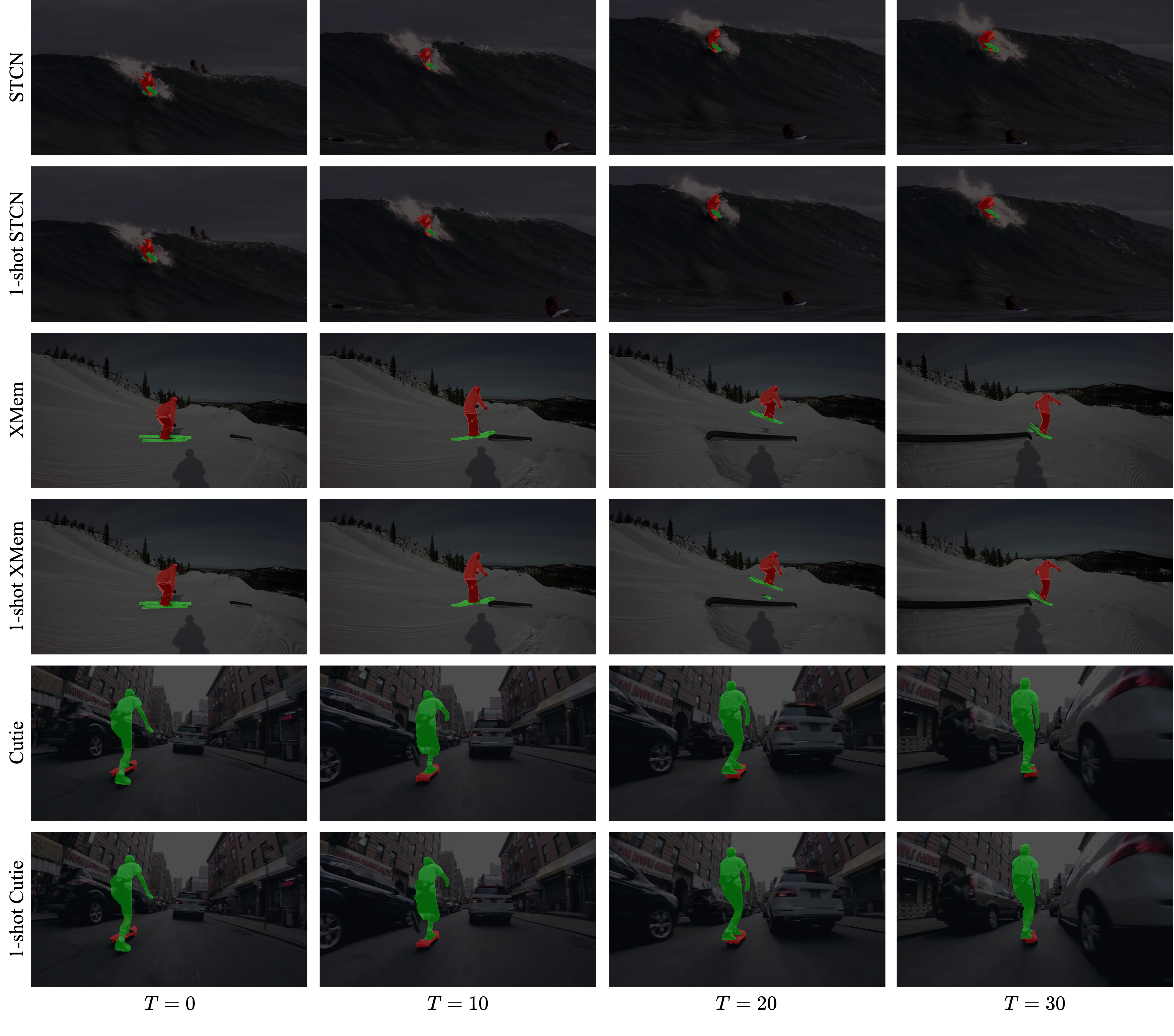}
    \caption{Qualitative comparisons of STCN~\cite{STCN}, XMem~\cite{Xmem}, and Cutie~\cite{Cutie} trained with one-shot training and fully-supervised training on YouTube-VOS 2018 \emph{val}. Although subtle errors can be observed in the fourth row, our predictions look satisfactory.}
    \label{fig:y18-vis}
\end{figure}

\paragraph{Comparisons on YouTube-VOS benchmarks.} YouTube-VOS 2018 \emph{val} and YouTube-VOS 2019 are more complicated and comprehensive benhcmarks for VOS than DAVIS benchmarks, which consist 91 unique objects out of 474 and 507 sequences, respectively. We employ our One-shot Training paradigm to STCN, XMem, and Cutie as done above, and compare them with the fully-supervised ones. In Table~\ref{tab:ytvos_res}, our approach reveals promising results by merely using $3.7\%$ labels of YouTube-VOS benchmarks (1 labeled frame per training video). In particular, our approach exhibits competitiveness with previous VOS solutions, such as STM~\cite{STM} and CFBI~\cite{CFBI}. Although there is still a gap between our approach and the fully supervised state-of-the-arts, our method significantly reduces the repetitive task of labeling.

\subsection{Analysis}
\label{sec:analysis}


\begin{figure}[ht]
    \centering
    \includegraphics[width=\linewidth]{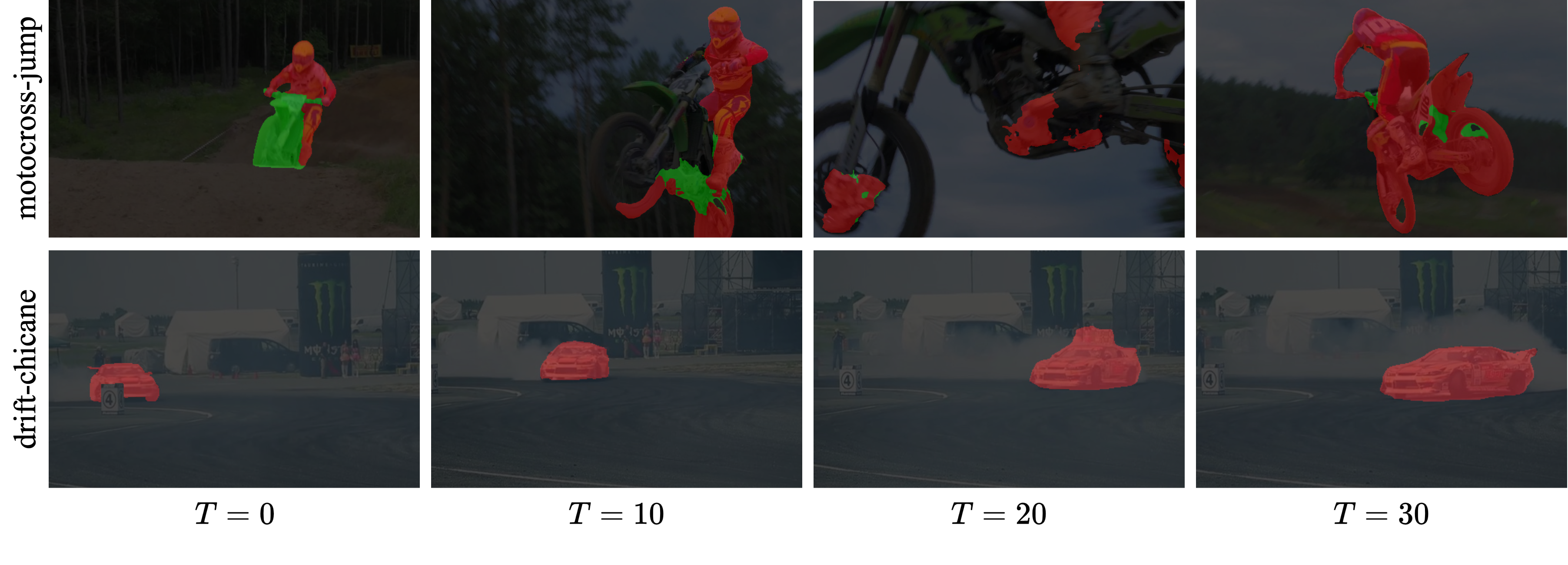}
    \caption{Visualization of coarse masks predicted by Cutie~\cite{Cutie} guided by an empty reference mask after 15,000~iterations. Predictions on the \emph{motocross-jump} sequence roughly capture objects across frames. In particular, results on the \emph{drift-chicane} are surprisingly coherent over time.}
    \label{fig:intrinsic}
\end{figure}

\paragraph{Intrinsic of One-shot Training.} To clearer illustrate the principle under our One-shot Training, we visualize the coarse predictions obtained from Cutie~\cite{Cutie} in Fig.~\ref{fig:intrinsic}, which is trained under the empty reference mask settings. Specifically, our One-shot Training stems from the observation that coarse masks can be obtained from a VOS network trained from a noisy input reference. Following Fig.~\ref{fig:key_concept}, we start with an empty object mask and ask the VOS network to predict the mask of another frame associated with a label. By aligning the prediction to the label, the VOS network learn to generate coarse masks, as shown in Fig.~\ref{fig:intrinsic}. Particularly, if we replace the empty object mask with the prediction mask after each iteration, the quality of input object mask will be gradually improved, resulting in beter prediction. Hence, a satisfactory VOS network can be trained using just one labeled frame, and for the sake of simplicity, we regard the first frame of a video as the annotated one. 

\begin{figure}[h!]
    \centering
    \includegraphics[width=\linewidth]{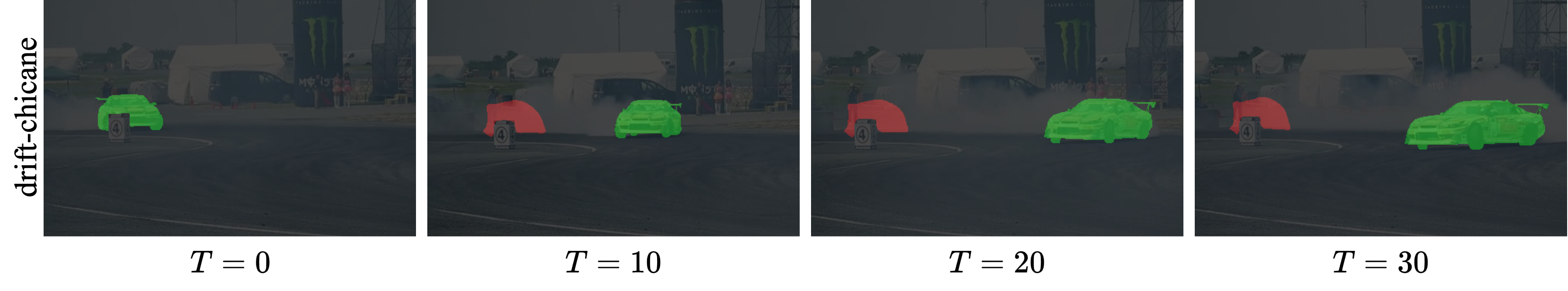}
    \caption{Cutie~\cite{Cutie} trained with \emph{T-step backward} One-shot Training from scratch overfits the \emph{drift-chicane} sequence of DAVIS 2017 \emph{val}. The groundtruth is in green while the red one is predictions. It shows that the network simply copies the initial mask for the predictions of the subseqeunt frames.}
    \label{fig:overfitting}
\end{figure}

\paragraph{Number of steps.} To thoughtfully analysis the impacts of \emph{T-step backward}, \emph{2-step backward} and \emph{1-step backward} discussed in Sec.~\ref{sec:backward-reconstruction}, we use the proposed One-shot Training method equipped with different backward patterns to train VOS networks without pretraining,~respectively. Table~\ref{tab:num_steps} shows that \emph{2-step backward} performs coherently better than \emph{T-step backward} when training from scratch. The reason lies in that traversing back with more steps increases the complexity of the VOS network, leading to the overfitting issue. We can observe this easily in Fig.~\ref{fig:overfitting}, in which the predictions for subsequent frames are almost identical to the initial masks of sequences.

\paragraph{The impacts of pretraining.} In Table~\ref{tab:davis_res}, we report results pretrained on the static image datasets following prior practices \cite{RGMP,STM,STCN}. As illustrated above, \emph{T-step backward} benefits from the pretraining. Hence, we train models with \emph{2-step backward} to explicitly investigate the role of pretraining in our approach. Table~\ref{tab:whether_using_pretraining} shows that the impact of pretraining is minimal, and our method is not heavily dependent on it. Specifically, the 1-shot XMem without pretraining performs slightly better than the pretrained one.

\begin{table}[t]
    \centering
    \begin{minipage}[t]{0.45\linewidth}
    \caption{Comparison of whether incorperating One-shot Training with synthetic video pretraining on DAVIS 2017 \emph{val}. All models are equipped with \textbf{2-step backward}.}
    \label{tab:whether_using_pretraining}
    \centering
    \begin{tabular}{lccc}
        \toprule
         Model  & $\mathcal{J}\&\mathcal{F}$ & $\mathcal{J}$ & $\mathcal{F}$ \\
         \midrule
         \multicolumn{4}{l}{\footnotesize \emph{with pretraining:}} \\[1.37mm]
         1-shot STCN~\cite{STCN}   & 81.7 & 78.6 & 84.8 \\ [1.5mm]
         1-shot XMem~\cite{Xmem}   & 82.4 & 79.2 & 85.5 \\ [1.5mm]
         1-shot Cutie~\cite{Cutie}  & 84.2 & 81.2 & 87.2 \\
         \midrule
         \multicolumn{4}{l}{\footnotesize \emph{without pretraining:}} \\[1.35mm]
         1-shot STCN~\cite{STCN}   & 78.0 & 75.2 & 80.9 \\ [1.5mm]
         1-shot XMem~\cite{Xmem}   & 82.7 & 79.4 & 86.1 \\ [1.5mm]
         1-shot Cutie~\cite{Cutie}  & 83.8 & 80.5 & 87.0 \\
         \bottomrule
    \end{tabular}
    \end{minipage}
    \hspace{0.05\linewidth}
    \begin{minipage}[t]{0.45\linewidth}
            \centering
            \caption{Comparison of different backward strategies of One-shot Training on DAVIS 2017 \emph{val}. All models are trained \textbf{from scratch} (without pretraining).}
            \label{tab:num_steps}
            \begin{tabular}{lccc}
                \toprule
                Model & $\mathcal{J}\&\mathcal{F}$ & $\mathcal{J}$ & $\mathcal{F}$ \\
                \midrule
                \multicolumn{4}{l}{\footnotesize \emph{T-step backward:}} \\[1mm]
                1-shot XMem~\cite{Xmem}  & 79.2 & 75.9 & 82.5\\
                1-shot Cutie~\cite{Cutie}  & 72.4 & 69.4 & 75.4 \\
                \midrule
                \multicolumn{4}{l}{\footnotesize \emph{2-step backward:}} \\[1mm]
                1-shot STCN~\cite{STCN}   & 78.0 & 75.2 & 80.9 \\
                1-shot XMem~\cite{Xmem}   & 82.7 & 79.4 & 86.1 \\
                1-shot Cutie~\cite{Cutie}  & 83.8 & 80.5 & 87.0 \\
                \midrule
                \multicolumn{4}{l}{\footnotesize \emph{1-step backward:}} \\[1mm]
                1-shot STCN~\cite{STCN}   & 77.5 & 74.8 & 80.2 \\
                \bottomrule
            \end{tabular}
    \end{minipage}
\end{table}

\paragraph{Qualitative results.} To gain a clearer insight of our approach, we visualize the predictions of STCN~\cite{STCN}, XMem~\cite{Xmem} and Cutie~\cite{Cutie} trained with our One-shot Training on DAVIS 2017 \emph{val} in Fig.~\ref{fig:d17-vis}. The predictions from the fully-supervised counterparts are also presented for reference. For each pair of comparisons, we randomly sample 1 sequence from DAVIS 2017 \emph{val}. On the third and fourth rows of Fig.~\ref{fig:d17-vis}, a failure case can be observed that our 1-shot model cannot distinguish the feets from the vehicle, as the feets are obscured by the vehicle. However, the overall differences between 1-shot models and their counterparts are hard to perceive. Fig.~\ref{fig:y18-vis} compares the predictions of our approach with the fully-supervised ones on YouTube-VOS 2018 \emph{val}. And it consistently shows that training VOS models in one-shot training paradigm is effective and it is extremely efficient since only one labeled frame per training sequence is required.

\section{Limitations} \label{limitations}

To distinguish the proposed One-shot Training from existing semi-supervised training strategies, we merely train VOS networks in a single-phase fashion without incorperating other semi-supervised technologies such as pseudo-labeling, leading to a suboptimal solution. However, we expect the simplicity of our approach can inspire more advanced semi-supervised training paradigm.

\section{Conclusion} \label{conclusion}

We propose a general One-shot Training framework for video object segementation that is remarkably simple yet highly effective, requiring only  a single-labeled frame per training video. Additionally, our approach is model-agnostic and demonstrate strong generalizability across various state-of-the-art VOS networks. Our approach utilizes a bi-directional training paradigm: during the time-forwrad inference, we start with a labeled initial frame and employ a VOS network to infer object masks sequencially. In the time-backward reconstruction, we initiate with the last inferred mask and go through the VOS network in reverse to predict the initial mask. Finally, the VOS network is updated by matching the prediction and the initial annotation. 

\newpage


{
	\bibliographystyle{ieee_fullname}
	\bibliography{one-shot-vos}

\begin{thebibliography}{10}

\bibitem{Xmem}
Ho~Kei Cheng and Alexander~G Schwing.
\newblock Xmem: Long-term video object segmentation with an atkinson-shiffrin
  memory model.
\newblock In {\em European Conference on Computer Vision}, pages 640--658.
  Springer, 2022.

\bibitem{Cutie}
Ho~Kei Cheng, Seoung~Wug Oh, Brian Price, Joon-Young Lee, and Alexander
  Schwing.
\newblock Putting the object back into video object segmentation.
\newblock {\em arXiv preprint arXiv:2310.12982}, 2023.

\bibitem{youtube-vos}
Ning Xu, Linjie Yang, Yuchen Fan, Dingcheng Yue, Yuchen Liang, Jianchao Yang,
  and Thomas Huang.
\newblock Youtube-vos: A large-scale video object segmentation benchmark.
\newblock {\em arXiv preprint arXiv:1809.03327}, 2018.

\bibitem{davis2016}
Federico Perazzi, Jordi Pont-Tuset, Brian McWilliams, Luc Van~Gool, Markus
  Gross, and Alexander Sorkine-Hornung.
\newblock A benchmark dataset and evaluation methodology for video object
  segmentation.
\newblock In {\em Proceedings of the IEEE/CVF Conference on Computer Vision and
  Pattern Recognition}, pages 724--732, 2016.

\bibitem{davis2017}
Jordi Pont-Tuset, Federico Perazzi, Sergi Caelles, Pablo Arbel{\'a}ez, Alex
  Sorkine-Hornung, and Luc Van~Gool.
\newblock The 2017 davis challenge on video object segmentation.
\newblock {\em arXiv preprint arXiv:1704.00675}, 2017.

\bibitem{Mast}
Zihang Lai, Erika Lu, and Weidi Xie.
\newblock Mast: A memory-augmented self-supervised tracker.
\newblock In {\em Proceedings of the IEEE/CVF Conference on Computer Vision and
  Pattern Recognition}, pages 6479--6488, 2020.

\bibitem{CRW}
Allan Jabri, Andrew Owens, and Alexei Efros.
\newblock Space-time correspondence as a contrastive random walk.
\newblock {\em Advances in neural information processing systems},
  33:19545--19560, 2020.

\bibitem{DUL}
Nikita Araslanov, Simone Schaub-Meyer, and Stefan Roth.
\newblock Dense unsupervised learning for video segmentation.
\newblock {\em Advances in neural information processing systems},
  34:25308--25319, 2021.

\bibitem{two-shot-vos}
Kun Yan, Xiao Li, Fangyun Wei, Jinglu Wang, Chenbin Zhang, Ping Wang, and Yan
  Lu.
\newblock Two-shot video object segmentation.
\newblock In {\em Proceedings of the IEEE/CVF Conference on Computer Vision and
  Pattern Recognition}, pages 2257--2267, 2023.

\bibitem{one-shot-vos}
Sergi Caelles, Kevis-Kokitsi Maninis, Jordi Pont-Tuset, Laura Leal-Taix{\'e},
  Daniel Cremers, and Luc Van~Gool.
\newblock One-shot video object segmentation.
\newblock In {\em Proceedings of the IEEE/CVF Conference on Computer Vision and
  Pattern Recognition}, pages 221--230, 2017.

\bibitem{static_images}
Federico Perazzi, Anna Khoreva, Rodrigo Benenson, Bernt Schiele, and Alexander
  Sorkine-Hornung.
\newblock Learning video object segmentation from static images.
\newblock In {\em Proceedings of the IEEE/CVF Conference on Computer Vision and
  Pattern Recognition}, pages 2663--2672, 2017.

\bibitem{OAVOS}
Paul Voigtlaender and Bastian Leibe.
\newblock Online adaptation of convolutional neural networks for video object
  segmentation.
\newblock {\em arXiv preprint arXiv:1706.09364}, 2017.

\bibitem{wotemporal}
K-K Maninis, Sergi Caelles, Yuhua Chen, Jordi Pont-Tuset, Laura Leal-Taix{\'e},
  Daniel Cremers, and Luc Van~Gool.
\newblock Video object segmentation without temporal information.
\newblock {\em IEEE transactions on pattern analysis and machine intelligence},
  41(6):1515--1530, 2018.

\bibitem{fastandrobust}
Andreas Robinson, Felix~Jaremo Lawin, Martin Danelljan, Fahad~Shahbaz Khan, and
  Michael Felsberg.
\newblock Learning fast and robust target models for video object segmentation.
\newblock In {\em Proceedings of the IEEE/CVF conference on computer vision and
  pattern recognition}, pages 7406--7415, 2020.

\bibitem{STM}
Seoung~Wug Oh, Joon-Young Lee, Ning Xu, and Seon~Joo Kim.
\newblock Video object segmentation using space-time memory networks.
\newblock In {\em Proceedings of the IEEE/CVF Conference on Computer Vision and
  Pattern Recognition}, pages 9226--9235, 2019.

\bibitem{aot}
Zongxin Yang, Yunchao Wei, and Yi~Yang.
\newblock Associating objects with transformers for video object segmentation.
\newblock {\em Advances in Neural Information Processing Systems},
  34:2491--2502, 2021.

\bibitem{xmempp}
Maksym Bekuzarov, Ariana Bermudez, Joon-Young Lee, and Hao Li.
\newblock Xmem++: Production-level video segmentation from few annotated
  frames.
\newblock In {\em Proceedings of the IEEE/CVF International Conference on
  Computer Vision}, pages 635--644, 2023.

\bibitem{state-aware-tracker}
Xi~Chen, Zuoxin Li, Ye~Yuan, Gang Yu, Jianxin Shen, and Donglian Qi.
\newblock State-aware tracker for real-time video object segmentation.
\newblock In {\em Proceedings of the IEEE/CVF Conference on Computer Vision and
  Pattern Recognition}, pages 9384--9393, 2020.

\bibitem{dyenet}
Xiaoxiao Li and Chen~Change Loy.
\newblock Video object segmentation with joint re-identification and
  attention-aware mask propagation.
\newblock In {\em Proceedings of the European Conference on Computer Vision},
  pages 90--105, 2018.

\bibitem{RGMP}
Seoung~Wug Oh, Joon-Young Lee, Kalyan Sunkavalli, and Seon~Joo Kim.
\newblock Fast video object segmentation by reference-guided mask propagation.
\newblock In {\em Proceedings of the IEEE/CVF International Conference on
  Computer Vision}, pages 7376--7385, 2018.

\bibitem{STCN}
Ho~Kei Cheng, Yu-Wing Tai, and Chi-Keung Tang.
\newblock Rethinking space-time networks with improved memory coverage for
  efficient video object segmentation.
\newblock {\em Advances in Neural Information Processing Systems},
  34:11781--11794, 2021.

\bibitem{colorizing}
Carl Vondrick, Abhinav Shrivastava, Alireza Fathi, Sergio Guadarrama, and Kevin
  Murphy.
\newblock Tracking emerges by colorizing videos.
\newblock In {\em Proceedings of the European Conference on Computer Vision},
  pages 391--408, 2018.

\bibitem{cycle-consist}
Xiaolong Wang, Allan Jabri, and Alexei~A Efros.
\newblock Learning correspondence from the cycle-consistency of time.
\newblock In {\em Proceedings of the IEEE/CVF Conference on Computer Vision and
  Pattern Recognition}, pages 2566--2576, 2019.

\bibitem{rethink_corr_learn}
Jiarui Xu and Xiaolong Wang.
\newblock Rethinking self-supervised correspondence learning: A video
  frame-level similarity perspective.
\newblock In {\em Proceedings of the IEEE/CVF International Conference on
  Computer Vision}, pages 10075--10085, 2021.

\bibitem{pseduo_label}
Yves Grandvalet and Yoshua Bengio.
\newblock Semi-supervised learning by entropy minimization.
\newblock {\em Advances in neural information processing systems}, 17, 2004.

\bibitem{lee2013pseudo}
Dong-Hyun Lee et~al.
\newblock Pseudo-label: The simple and efficient semi-supervised learning
  method for deep neural networks.
\newblock In {\em Workshop on challenges in representation learning, ICML},
  volume~3, page 896, 2013.

\bibitem{osmn}
Linjie Yang, Yanran Wang, Xuehan Xiong, Jianchao Yang, and Aggelos~K
  Katsaggelos.
\newblock Efficient video object segmentation via network modulation.
\newblock In {\em Proceedings of the IEEE/CVF International Conference on
  Computer Vision}, pages 6499--6507, 2018.

\bibitem{feelvos}
Paul Voigtlaender, Yuning Chai, Florian Schroff, Hartwig Adam, Bastian Leibe,
  and Liang-Chieh Chen.
\newblock Feelvos: Fast end-to-end embedding learning for video object
  segmentation.
\newblock In {\em Proceedings of the IEEE/CVF International Conference on
  Computer Vision}, pages 9481--9490, 2019.

\bibitem{trackseg}
Xi~Chen, Zuoxin Li, Ye~Yuan, Gang Yu, Jianxin Shen, and Donglian Qi.
\newblock State-aware tracker for real-time video object segmentation.
\newblock In {\em Proceedings of the IEEE/CVF International Conference on
  Computer Vision}, pages 9384--9393, 2020.

\bibitem{CFBI}
Zongxin Yang, Yunchao Wei, and Yi~Yang.
\newblock Collaborative video object segmentation by foreground-background
  integration.
\newblock In {\em Proceedings of the European Conference on Computer Vision},
  pages 332--348, 2020.

\bibitem{rdevos}
Mingxing Li, Li~Hu, Zhiwei Xiong, Bang Zhang, Pan Pan, and Dong Liu.
\newblock Recurrent dynamic embedding for video object segmentation.
\newblock In {\em Proceedings of the IEEE/CVF Conference on Computer Vision and
  Pattern Recognition}, pages 1332--1341, 2022.

\end{thebibliography}
}

\end{document}